\documentclass[acmsmall,screen,nonacm]{acmart}

\AtBeginDocument{%
  }

\usepackage{tcolorbox}
\usepackage{multirow}
\usepackage{booktabs}
\usepackage{enumitem}
\usepackage{xcolor}
\usepackage{caption}
\usepackage{subcaption}
\usepackage{wrapfig}
\setcopyright{acmcopyright}
\copyrightyear{2024}
\acmYear{2024}
\acmDOI{XXXXXXX.XXXXXXX}
\acmConference[FSE]{2024}{Galinas, Brazil}

\acmBooktitle{}

\newcommand{\tool}{\texttt{VeriStable}}
\newcommand{\crown}{\texttt{$\alpha$-$\beta$-CROWN}}
\newcommand{\nnenum}{\texttt{nnenum}}
\newcommand{\marabou}{\texttt{Marabou}}
\newcommand{\eran}{\texttt{ERAN}}
\newcommand{\reluplex}{\texttt{Reluplex}}
\newcommand{\reluval}{\texttt{Reluval}}
\newcommand{\neurify}{\texttt{Neurify}}
\newcommand{\nnv}{\texttt{NNV}}
\newcommand{\dnnv}{\texttt{DNNV}}
\newcommand{\verinet}{\texttt{VeriNet}}
\newcommand{\mnbab}{\texttt{MN-BaB}}

\newcommand{\vnncomp}{VNN-COMP'22}
\newcommand{\planet}{\texttt{Planet}}
\newcommand{\neuralsat}{\texttt{NeuralSAT}}

\usepackage[vlined,figure,linesnumbered]{algorithm2e}

\SetCommentSty{haicommentfont}

\newcommand{\functiontextformat}[1]{\textrm{\texttt{#1}}}

\newtoggle{usecomment}
\settoggle{usecomment}{true}

\newcommand{\tvn}[1]{\iftoggle{usecomment}{{\color{red}{[TVN]: #1}}}{}}

\newcommand{\hd}[1]{\iftoggle{usecomment}{{\color{blue}{[HD]: #1}}}{}}

\newcommand{\ignore}[1]{}

\usepackage{makecell}
\usepackage{tcolorbox}

\SetKwFor{Parfor}{parfor}{do}{endparfor}

\SetKwInOut{Input}{input}
\SetKwInOut{Output}{output}
\SetKw{Break}{break}
\SetKw{Continue}{continue}
\SetKwFunction{Backtrack}{Backtrack}
\SetKwFunction{Decide}{Decide}
\SetKwFunction{BCP}{BCP}
\SetKwFunction{Deduce}{Deduce}
\SetKwFunction{AnalyzeConflict}{AnalyzeConflict}
\SetKwFunction{BooleanAbstraction}{BooleanAbstraction}
\SetKwFunction{Restart}{Restart}
\SetKwFunction{Select}{Select}
\SetKwFunction{Add}{Add}
\SetKwFunction{Sort}{Sort}
\SetKwFunction{Stabilize}{Stabilize}
\SetKwFunction{isVerified}{isVerified}
\SetKwFunction{isComplete}{isComplete}
\SetKwFunction{MIP}{CreateMILP}
\SetKwFunction{Select}{Select}
\SetKwFunction{GetUnassignedVariable}{GetUnassignedVariable}
\SetKwFunction{Maximize}{Maximize}
\SetKwFunction{Minimize}{Minimize}
\SetKwFunction{isStable}{isStable}
\SetKwFunction{StabilizeCondition}{StabilizeCondition}

\SetKwData{implicationgraph}{igraph}
\SetKwData{conflict}{conflict}
\SetKwData{none}{none}
\SetKwData{infeasible}{INFEASIBLE}
\SetKwData{unreachable}{UNREACHABLE}
\SetKwData{assignment}{$\sigma$}
\SetKwData{network}{$\alpha$}
\SetKwData{dl}{dl}
\SetKwData{clauses}{clauses}
\SetKwData{clause}{clause}
\SetKwData{igraph}{igraph}
\SetKwData{cex}{cex}
\SetKwData{sat}{sat}
\SetKwData{unsat}{unsat}
\SetKwData{status}{status}
\SetKwData{nodes}{nodes}
\SetKwData{learnedclauses}{learned\_clauses}
\SetKwData{model}{model}
\SetKwData{assignments}{assignments}
\SetKwData{issat}{is\_sat}
\SetKwData{conflictclause}{conflict\_clause}
\SetKwData{isconflict}{is\_conflict}

\SetKwData{true}{true}
\SetKwData{false}{false}
\DontPrintSemicolon

\begin{document}

\title{Harnessing Neuron Stability to Improve DNN Verification}

\author{Hai Duong}
\email{hduong22@gmu.edu}
\affiliation{%
  \institution{George Mason University}
  \country{USA}
}

\author{Dong Xu}
\email{dx3yy@virginia.edu}
\affiliation{
\institution{University of Virginia}
\country{USA}
}

\author{ThanhVu Nguyen}
\email{tvn@gmu.edu}
\affiliation{%
  \institution{George Mason University}
  \country{USA}
}
\author{Matthew B. Dwyer}
\email{matthewbdwyer@virginia.edu}
\affiliation{
\institution{University of Virginia}
\country{USA}
}

\begin{abstract}
Deep Neural Networks (DNN) have emerged as an effective approach to tackling real-world problems.
However, like human-written software, DNNs are susceptible to bugs and attacks. This has generated significant interests in developing effective and scalable DNN verification techniques and tools.

Recent developments in DNN verification have highlighted the potential of constraint-solving approaches that combine abstraction techniques with SAT solving.  
Abstraction approaches are effective at precisely encode neuron behavior when it is linear, but they lead to overapproximation and combinatorial scaling when behavior is non-linear.
SAT approaches in DNN verification have incorporated standard DPLL techniques, but have overlooked important optimizations found in modern SAT solvers that help them scale on industrial benchmarks. 

In this paper, we present \tool{}, a novel extension of recently proposed DPLL-based constraint DNN verification approach. 
\tool{} leverages the insight that while neuron behavior may be non-linear across the entire DNN input space, at intermediate states computed during verification many neurons may be constrained to have linear behavior -- these neurons are stable.  Efficiently detecting stable neurons reduces combinatorial complexity without compromising the precision of abstractions.  Moreover, the structure of clauses arising in DNN verification problems shares important characteristics with industrial SAT benchmarks.  We adapt and incorporate multi-threading and restart optimizations targeting those characteristics to further optimize DPLL-based DNN verification.

We evaluate the effectiveness of \tool{} across a range of challenging benchmarks including fully-connected feedforward networks (FNNs), convolutional neural networks (CNNs) and residual networks (ResNets) applied to the standard MNIST and CIFAR datasets.
Preliminary results show that \tool{} is competitive and outperforms state-of-the-art DNN verification tools, 
including \crown{} and \mnbab{}, 
the first and second performers of the VNN-COMP, respectively. 

\end{abstract}




\maketitle




\ignore{
\section*{TBDs}
\begin{itemize}
\item inconsistent use of text formatting for function names (textrm vs textsc) (sec 4.1)
\end{itemize}
}

\section{Introduction}
Increasingly deep neural networks (DNN) are being employed as components of
mission-critical systems across a range of application domains, such as
autonomous driving~\cite{lee2023end,shao2023safety},
medicine~\cite{bizjak2022deep,morris2023deep},
and infrastructure monitoring~\cite{ewald2022perception,ye2023deep}.
DNNs require high levels of assurance in order to confidently deploy
them in such systems.

As with traditional software, testing DNNs using rigorous coverage
criteria~\cite{surpriseadequacy,neuralcoverage,deephyperion,IDC} is
necessary but not sufficient for critical deployments.
To provide further evidence that DNN behavior meets expectations
researchers have developed a range of techniques for verifying
specifications formulated as pre/post-condition specifications
that can be rendered in a canonical form~\cite{shriver2021reducing}.
Many dozens of DNN verifiers have been reported in the
literature and a yearly competition has documented advances in the capabilities of such techniques~\cite{bak2021second,muller2022third,brix2023first}.

Despite those advances, as with traditional software verification, DNN verification suffers from exponential worst-case complexity~\cite{katz2017reluplex}.
To understand why, consider the common case of DNNs with neurons using the rectified linear unit (ReLU) activation function~\cite{goodfellow2016deep}.
The input of a neuron is defined as the weighted sum of the outputs of neurons preceding it in the computation graph, where the weights are the learned parameters of the DNN.
The output of a neuron applies the ReLU function, $ReLu(x) = max(x,0)$, to its input.
This can be encoded as the disjunction of two partial linear functions -- the zero and identity functions defined over negative and non-negative domains, respectively.
When a neuron's input is positive, $x > 0$, the neuron is said to be \textit{active}; otherwise, it is \textit{inactive}.
For a given input, running inference on a DNN causes each neuron to be either active or inactive.
The vector of Boolean values representing each neuron's activation status is called an \textit{activation pattern} for the input.
In the worst-case, if the DNN has $n$ neurons then there are $2^n$ activation patterns.
Realistic DNNs, like ResNet~\cite{ResNetv1}, can have 10s of thousands of
neurons making it extremely challenging to reason about the full space of activation patterns.

While this complexity seems daunting, history has shown that despite
the worst-case exponential growth of verification problems, like
propositional satisfiability (SAT)~\cite{cook1971complexity}, it is possible to solve very large
problem instances with sophisticated algorithmic techniques~\cite{biere2009handbook}.
Modern SAT solvers aim to determine if there exists an assignment of truth
values to propositional variables that satisfies a given set of logical constraints.
They are based on the classic Davis-Putnam-Logemann-Loveland (DPLL) algorithm~\cite{davis1962machine} which searches the space of assignments by alternating between \textit{deciding} how to extend a partial assignment -- by choosing a variable and a truth value for it -- and identify additional assignments that are \textit{implied} by that decision.
State-of-the-art solvers also incorporate a plethora of optimizations like Conflict-Driven Clause Learning (CDCL) to short-circuit later portions of the search~\cite{zhang2001efficient},
heuristics to restart search with learned clauses~\cite{biere2008picosat}, and parallel exploration of variable assignments~\cite{le2017painless}.  
Modern satisfiability modulo theory (SMT) solvers combine 
combine DPLL with theory-specific symbolic deduction methods that adapt and integrate with CDCL to form DPLL(T), where T stands for theory~\cite{nieuwenhuis2006solving}.

Most prior work on DNN verification either used SMT to discharge sub-problems formed by search of the space of activation patterns~\cite{katz2017reluplex,katz2022reluplex,huang2017safety}, applied forms of abstract interpretation to approximate the disjunctive neuron behavior~\cite{singh2018fast, singh2019abstract,singh2018boosting,wang2018efficient,bak2021nnenum,singh2019beyond,wang2021beta,xu2020fast,gehr2018ai2,tjeng2017evaluating,henriksen2020efficient,botoeva2020efficient,ferrari2022complete}, or combined these approaches~\cite{katz2019marabou,ehlers2017formal}.

\begin{figure}[t]
    \begin{minipage}[b]{\textwidth}
    \centering
        \begin{minipage}[t]{0.33\textwidth}
            \centering  
            \includegraphics[width=\linewidth]{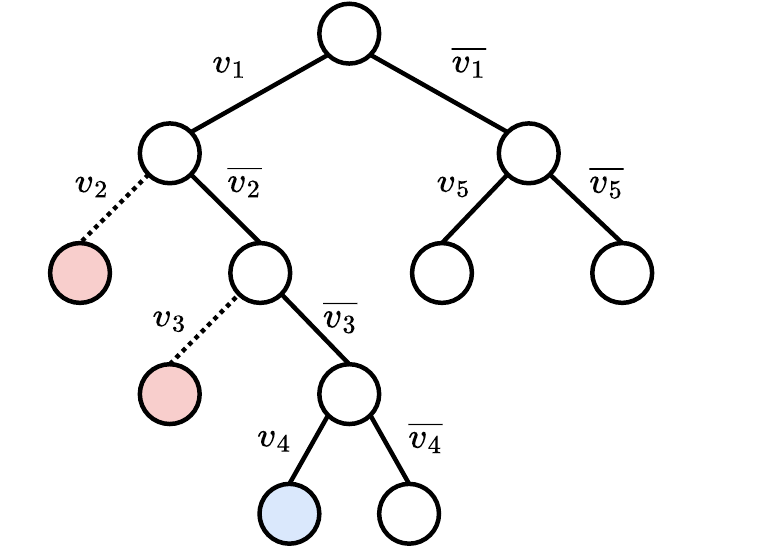}
            \caption*{(a) \neuralsat{}}
        \end{minipage}
        \hspace{1cm}
        \begin{minipage}[t]{0.3\textwidth}
            \centering
            \includegraphics[width=\linewidth]{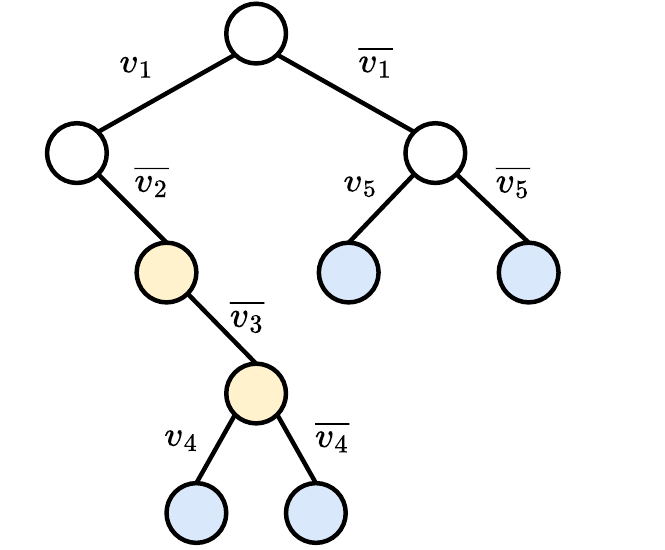}
            \caption*{(b) \tool{}}
        \end{minipage}
        \vspace{-0.1in}
        \caption{The tree of activation patterns computed by \neuralsat{} (left) and \tool{} (right) at corresponding points during a verification run.}
        \label{fig:tree}
    \end{minipage}
\end{figure}

The success of these methods inspired recent work that adapts DPLL(T) to DNN verification by incorporating an abstraction-based theory solver~\cite{duong2023dpllt} to realize the \neuralsat{} verifier.
In \neuralsat{}, propositional variables encode whether a neuron is
active or inactive, and additional constraints encode the weighted sums for each neuron input.  As illustrated on the left of Fig.~\ref{fig:tree}, \neuralsat{} searches the space of activation patterns for a DNN; here $v_i$ and
$\overline{v_i}$ denote that the $i$th neuron is active or inactive, respectively, and a path in the tree is a partial activation pattern.
As we discuss in \S\ref{sec:background}, \neuralsat{}'s contribution lies in combining DPLL(T) with a custom theory solver, that uses abstraction, to determine whether a partial activation pattern implies the specified property or implies conflict clauses that can prune subsequent search through CDCL.

In this paper, we further extend DPLL(T)-based DNN verification in two
significant ways.

First, we propose a method for computing, from a partial activation pattern, a set of neurons that must be either active or inactive -- such a neuron is said to be \textit{stable}.  
Stable neurons eliminate the need for deciding their activation status later in the search and thereby lead to combinatorial reduction in the search.  
Unlike prior work~\cite{xiao2018training,crownIBP,li2022can,chen2022linearity} which seeks to modify the network to create neurons that are stable for inputs described by the specification precondition, our approach (1) does not modify the network being verified and (2) detects neurons that are stable relative to subsets of the precondition.
Our method can be thought of as \textit{state-sensitive neuron stabilization}, where the state is a partial activation pattern encoding a subset of the precondition.
Fig.~\ref{fig:tree} depicts how after $v_1$ is decided our method, \tool{}, stabilizes two neurons to be stable and inactive -- shown in yellow -- which eliminates the need to search their active branches -- shown in red -- as required by \neuralsat{}.  In this depiction, $v_1$ constitutes the state relative to which $v_2$ and $v_3$ are determined to be stable.

Second, we adapt parallelization techniques and restart heuristics from propositional SAT solvers to target the problem of DNN verification.  Fig.~\ref{fig:tree} depicts how \neuralsat{}'s search frontier is a single state -- shown in blue -- and how \tool{} can expand a broader frontier and do so in parallel.  As depicted, stabilization and parallelization are synergistic in that the former reduces the tree width which allows the latter to process a larger percentage of the tree.

While we developed these methods in the context of DPLL(T), these conceptual contributions are broadly applicable to any DNN verification approach that performs a search of the space of activation patterns and splits the search based on the activation status of neurons, such as~\cite{wang2021beta,bak2021nnenum}.
We implement the methods in \tool{} and demonstrate empirically that each of the methods it incorporates leads it to outperform \neuralsat{}, that in combination all of the methods lead to a 12-fold increase in the ability to solve verification problems, and that it establishes a new state-of-the-art in DNN verification compared with the top performers in the most recent DNN verifier competition~\cite{muller2022third}.

The key contributions of the paper lie in:
\begin{itemize}
\item developing a novel approach that computes state-sensitive neuron stability to eliminate the need for neuron splitting in DNN verification;
\item adaptation of advanced SAT optimizations into a DPLL(T)-based verification algorithm;
\item evaluation results using a new challenging DNN verification benchmark, as well as existing benchmarks, that demonstrate a 12-fold improvement in performance and that \tool{} establishes the state-of-the-art in DNN verifier performance; and
\item release of an open source implementation of \tool{}\footnote{\url{https://github.com/veristable/veristable}} accepting verification problems in standard formats to promote the application of DNN verification and comparative evaluation.
\end{itemize}

\section{Background}\label{sec:background}

\subsection{The DNN verification problem}\label{sec:nnverif}

A \emph{neural network} (\textbf{NN})~\cite{Goodfellow-et-al-2016} consists of an input layer, multiple hidden layers, and an output layer. Each layer contains neurons connected to neurons in previous layers via predefined weights obtained through training with data. A \emph{deep} neural network (\textbf{DNN}) is an NN with at least two hidden layers.

The output of a DNN is computed by iteratively calculating the values of neurons in each layer. Neurons in the input layer receive the input data. Neurons in the hidden layers compute their values through an \emph{affine transformation} followed by an \emph{activation function}, like the popular Rectified Linear Unit (ReLU) activation.

For ReLU activation, the value of a hidden neuron \(y\) is given by $ReLU(w_1v_1 + \dots{} + w_nv_n + b)$, where \(b\) is the bias parameter for \(y\), \(w_i, \dots, w_n\) are the weights of \(y\), \(v_1,\dots,v_n\) are the neuron values from the preceding layer, \(w_1v_1 + \dots + w_nv_n+b\) represents the affine transformation, and \(ReLU(x) = \max(x,0)\) defines the ReLU activation. 
A ReLU-activated neuron is said to be \emph{active} if its input value is greater than zero and \emph{inactive} otherwise.

We note that ReLU DNNs are popular because they tend to be sparsely activated~\cite{glorot2011deep} and $\max(x,0)$ is efficient to compute which leads to efficient training and inference.
Moreover, they avoid the vanishing gradient problem~\cite{goodfellow2016deep} which speeds training convergence especially in deep networks.
This makes ReLU networks an important class to target, but we note that \tool{} applies equally well to DNNs using other piecewise-linear activation functions, such as leaky ReLU~\cite{maas2013rectifier} and parametric ReLU~\cite{he2015delving}.




\paragraph{DNN Verification} Given a DNN \(N\) and a property $\phi$, the \emph{DNN verification problem} asks if $\phi$ is a valid property of $N$.
Typically, $\phi$ is a formula of the form $\phi_{in} \Rightarrow \phi_{out}$, where $\phi_{in}$ is a property over the inputs of $N$ and $\phi_{out}$ is a property over the outputs of $N$.
This form of property has been used to encode safety and security requirements of DNNs, e.g., safety specifications to avoid collision in unmanned aircraft~\cite{kochenderfer2012next}.  
\ignore{It has been shown broad class of properties like those described above can by transformed into robustness properties~\cite{shriver2021reducing}, which state that 
small input perturbations lead to identical network outputs. 
}
A DNN verifier attempts to find a \emph{counterexample} input to $N$ that satisfies $\phi_{in}$ but violates $\phi_{out}$.  If no such counterexample exists, $\phi$ is a valid property of $N$. Otherwise, $\phi$ is not valid and the counterexample can be used to retrain or debug the DNN~\cite{huang2017safety}.

\begin{wrapfigure}{r}{0.50\textwidth}
    \vspace{-0.1in}
    \centering
    \includegraphics[width=1\linewidth]{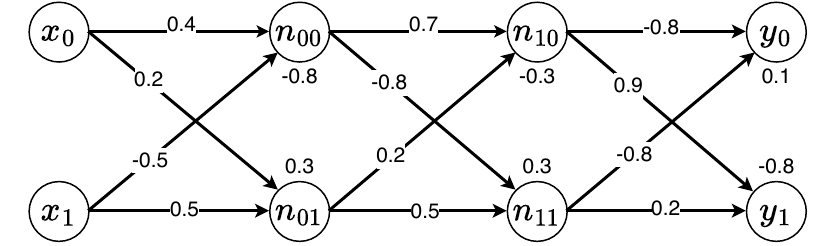}
    \caption{An FNN with ReLU.}\label{fig:example}
    \vspace{-0.1in}
\end{wrapfigure}
\paragraph{Example} 
Fig.~\ref{fig:example} shows a simple DNN with two inputs $x_0,x_1$, four hidden neurons $n_{00},n_{01},n_{10},n_{11}$, and two outputs $y_0, y_1$. 
The weights of a neuron are shown on its incoming edges, and the bias is shown above or below each neuron. The outputs of the hidden neurons  are computed by the affine transformation and ReLU, 
e.g., $n_{00} = ReLU(0.4 x_0 - 0.5 x_1 - 0.8)$. The output neuron is computed with just the affine transformation, i.e., $y_0 = -0.8 n_{10} - 0.8 n_{11} + 0.1$.
A valid property for this DNN is that the output is $y_0 > y_1$ for any inputs $x_0 \in [-2.0, 2.0], x_ 1 \in [-1.0, 1.0]$.

ReLU-based DNN verification is NP-Complete~\cite{katz2017reluplex} and thus can be formulated as an SAT or SMT checking problem.
Direct application of SMT solvers does not scale to the large and complex formulae encoding real-world, complex DNNs.
While custom solvers, like \planet{} and \reluplex{}, retain the soundness, completeness,
and termination of SMT and improve on the performance of a direct SMT encoding, they do not scale to handle realistic DNNs~\cite{bak2021second}.

\ignore{
\begin{figure}
    \begin{minipage}[b]{1\linewidth}
        \centering
        \begin{minipage}[c]{0.25\linewidth}
            \includegraphics[width=\linewidth]{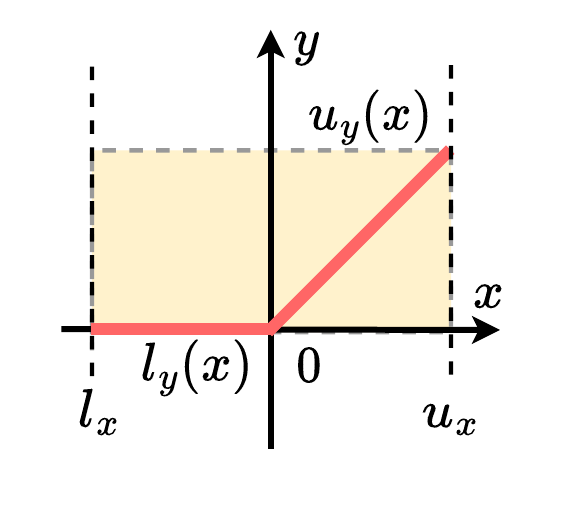}
            \vspace*{-10mm}
            \caption*{(a) interval}
        \end{minipage}
        \hspace{0.05\textwidth}%
        \begin{minipage}[c]{0.25\linewidth}
            \includegraphics[width=\linewidth]{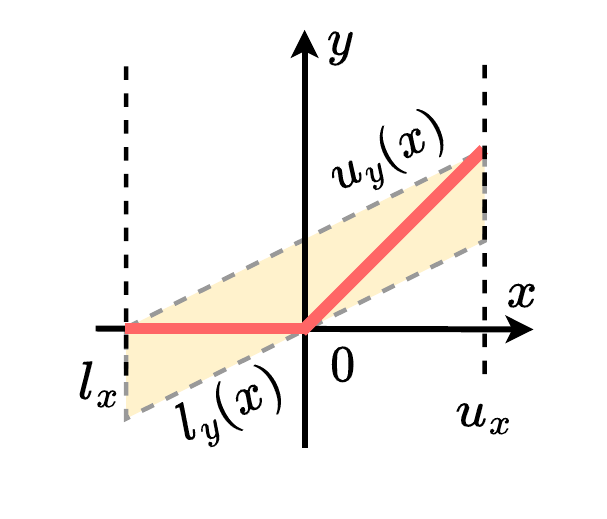}
            \vspace*{-10mm}
            \caption*{(b) zonotope}
        \end{minipage}
        \hspace{0.05\textwidth}%
        \begin{minipage}[c]{0.25\linewidth}
            \includegraphics[width=\linewidth]{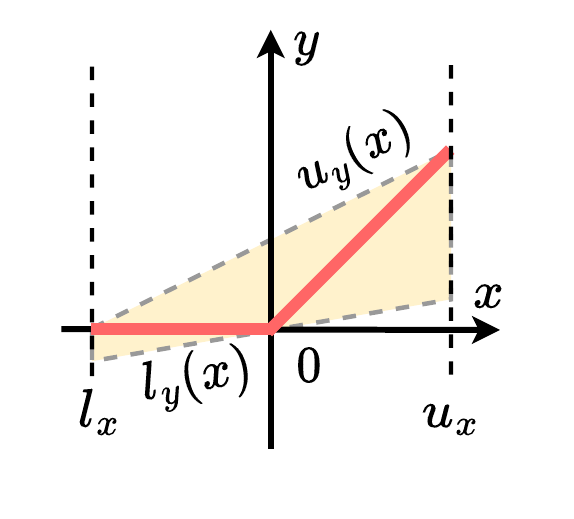}
            \vspace*{-10mm}
            \caption*{(c) polytope}
        \end{minipage}
        \vspace*{-3mm}
        \caption{Abstractions for ReLU: (a) interval, (b) zonotope, and (c) polytope. Notice that ReLU is a non-convex region (red line) while all abstractions are convex regions}\label{fig:abs}
    \end{minipage}
\end{figure}
}
\paragraph{Abstraction}
Applying techniques from abstract interpretation~\cite{cousot1977abstract},
abstraction-based DNN verifiers overapproximate nonlinear computations (e.g., ReLU) of the network using linear abstract domains such as intervals~\cite{wang2018formal} or polytopes~\cite{singh2019abstract,xu2020fast}. 
This allows abstraction-based DNN verifiers to side-step the disjunctive splitting that is the performance bottleneck
of constraint-based DNN verifiers.

\subsection{DPLL(T)-based DNN Verification}\label{sec:neuralsat}
\begin{wrapfigure}{r}{0.33\linewidth}
\centering
    \vspace*{-0.1in}
    \includegraphics[width=1\linewidth]{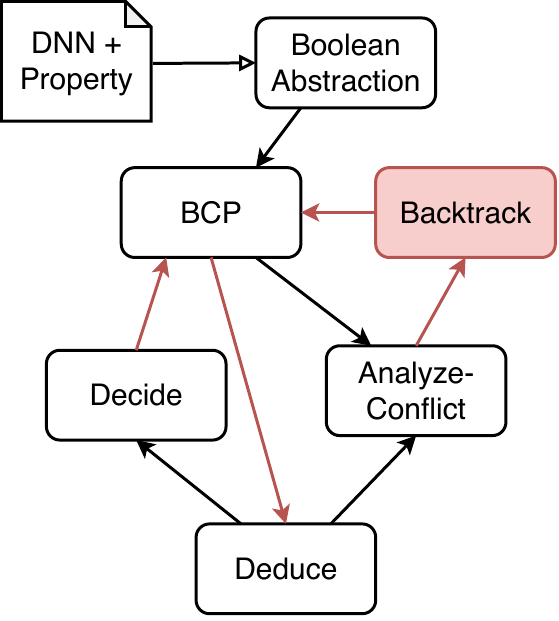}
    \caption{\neuralsat{} Architecture}\label{fig:neuralsat}
\end{wrapfigure}

While abstraction is crucial to the performance of DNN verification techniques,  recent work on \neuralsat{}~\cite{duong2023dpllt} shows that combining it with the DPLL(T) approach of modern SMT solvers~\cite{kroening2016decision,moura2008z3,barrett2011cvc4} can further improve the scalability of DNN verification.
Fig.~\ref{fig:neuralsat} gives an overview of \neuralsat{},  which consists of a theory solver (Deduce) and standard DPLL components (everything else).


\neuralsat{} constructs a propositional formula representing neuron activation status (Boolean Abstraction) and searches for satisfying truth assignments while employing a DNN-specific theory solver to check feasibility with respect to DNN constraints and properties. The process integrates standard DPLL components, which include Deciding variable assignments, and performing Boolean constraint propagation (BCP), with DNN-specific theory solving (Deduce), which uses LP solving and the polytope abstraction to check the satisfiability of assignments with the property of interest. If satisfiability is confirmed, it continues with new assignments; otherwise, it analyzes and learns conflict clauses (Analyze Conflict) to backtrack. \neuralsat{} continues it search until it either proves the property (\texttt{unsat}) or finds a total assignment (\texttt{sat}). In \S\ref{sec:dpll} we describe how these DPLL components are adapted and incorporated into \tool{}.

\subsection{Neuron Stability}
A ReLU neuron is \textit{stable} relative to a given specification when it is in either its active or inactive phase for all inputs satisfying the specification's precondition.
Researchers have observed that stable neurons have the potential to improve verifier performance, since they tend to linearize the otherwise highly non-linear computation encoded in the NN.
However, in prior work this required modifying the NN.
They have done this by increasing stability through a training objective~\cite{xiao2018training} or by identifying stable neurons and applying non-standard modifications that use strictly linear activation functions for those neurons~\cite{chen2022linearity, crownIBP}.  Importantly, this means that these techniques do not verify the original neural network.

We develop a method that can identify and exploit stable neurons while verifying the original network.
Moreover, we observe that a neuron may be stable relative to a subset of a specification's pre-condition.
Our method identifies when the verifier is analyzing such a subset which allows for a finer state-sensitive notion of neuron stability to be exploited.
In \S\ref{sec:stabilization} we define a method that encodes such subsets as partial activation patterns of neurons which allows neurons relative to that subset to be computed and subsequent verification to be more efficient.

\section{Overview and Illustration}
\ignore{
\begin{figure}
    \begin{minipage}[b]{1\textwidth}
        \centering
        \begin{minipage}[t]{0.35\linewidth}
            \includegraphics[width=0.75\linewidth]{figure/old.pdf}
            \caption*{(a) \neuralsat{}}
        \end{minipage}
        \hspace{1cm}
        \begin{minipage}[t]{0.35\linewidth}
            \includegraphics[width=\linewidth]{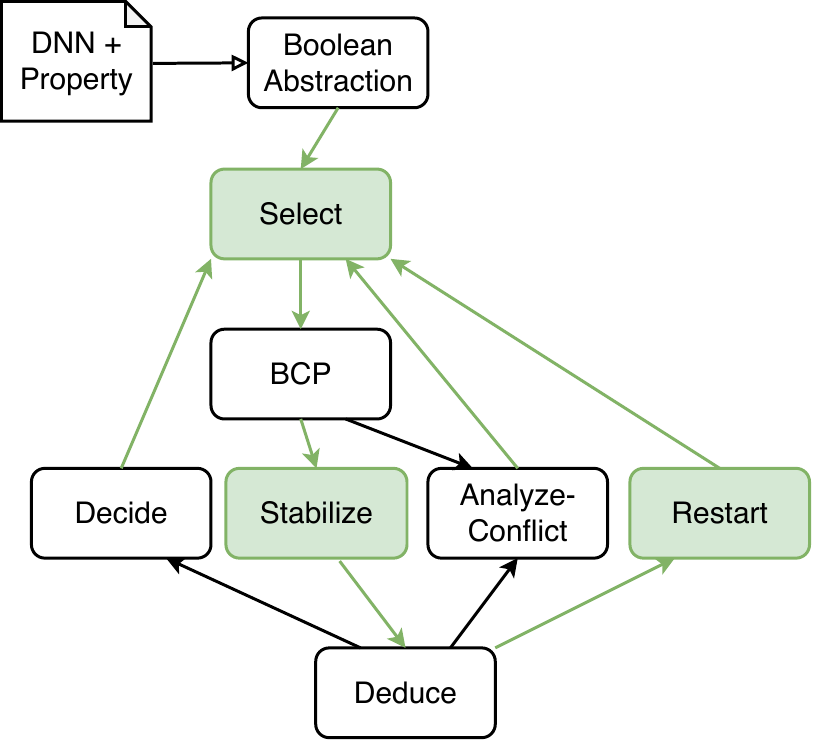}
            \caption*{(b) \tool{}}
        \end{minipage}
        \caption{\neuralsat{} vs. \tool{}. \tool{} adds three components (green) and removes Backtrack (red).}\label{fig:overview}
    \end{minipage}
\end{figure}
}
\subsection{Overview}\label{sec:overview}

Fig.~\ref{fig:overview} gives an overview of \tool{} DPLL(T) approach.
Compared to the \neuralsat{} DPLL(T) algorithm in Fig.~\ref{fig:neuralsat},  \tool{} also consists of standard DPLL components \functiontextformat{Decide}, \functiontextformat{BCP}, and \functiontextformat{AnalyzeConflict}, and the DNN-dedicated theory solver for \functiontextformat{Deduce}.
However, the \tool{} approach extends and significantly improves the performance of \neuralsat{} in three main ways. 

\begin{wrapfigure}{r}{0.41\linewidth}
    \vspace*{-0.1in}
    \centering
    \includegraphics[width=1\linewidth]{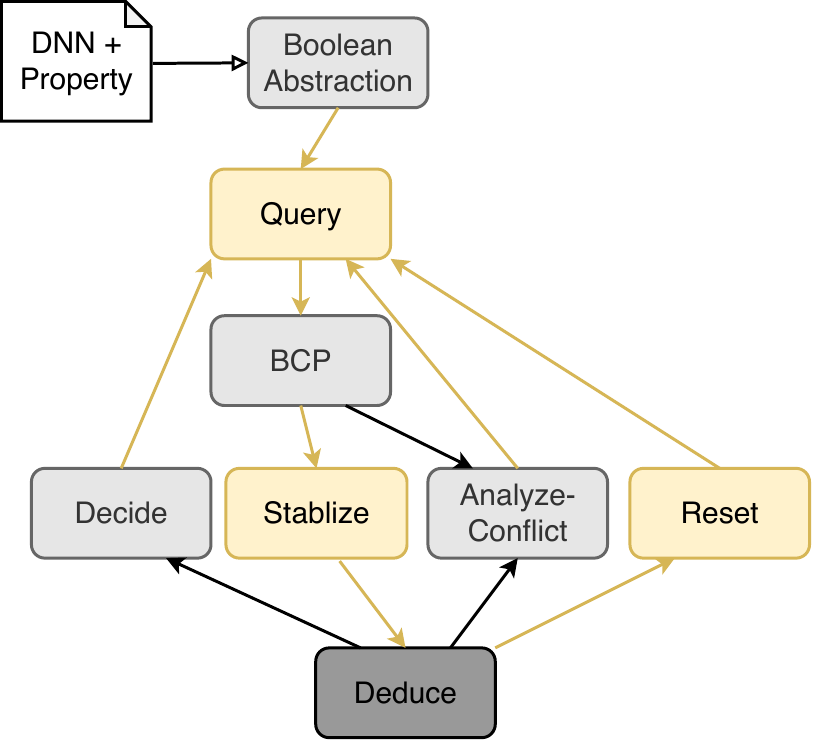}
    \caption{\tool{} Architecture}\label{fig:overview}
\end{wrapfigure}
First, for theory solving, we leverage the concept of \emph{neuron stability} to improve bound tightening and infer when neurons have linear behavior (Stabilize). This improves abstraction precision and eliminates the need to decide the activation status of neurons. Second, \tool{} employs a distributed search tree data structure to develop a parallel DPLL(T) approach. This allows \tool{} to leverage multicore processing and simultaneously analyze multiple possible assignments (Select).  This replaces Backtrack from DPLL(T) because it considers multiple branches simultaneously (including the one that would be backtracked to if run sequentially). Finally, \tool{} adopts \emph{restart heuristics} from modern SAT solving (e.g., PicoSAT~\cite{biere2008picosat}) to escape local optima (Restart). As we will discuss later, restarting especially benefits "hard" DNN problems by enabling better clause learning and exploring different decision orderings. 
In combination these techniques allow  \tool{} to solve 12 times more problems than \neuralsat{} (\S\ref{sec:eval}).

\subsection{Illustration}\label{sec:motiv}
We illustrate how \tool{} operates to verify the specification 
\[(x_0, x_1) \in [-2.0, 2.0] \times [-1.0, 1,0] \Rightarrow y_0 > y_1\]
on the DNN in Fig.~\ref{fig:example} by proving that the negation of the problem, $\alpha \land (-2.0 \le x_0 \le 2.0) \land (-1.0 \le x_1 \le 1.0) \land (y_0 \le y_1)$, where $\alpha$ is a logical formula encoding the DNN, is unsatisfiable.

Verification computes an interval approximating each neuron including
the outputs, $y_i \in [l_{y_i},u_{y_i}]$.  
So if $l_{y_0} > u_{y_1}$ then the post-condition $\phi_{out} = y_0 > y_1$. In our explanation  
$p = l_{y_0} - u_{y_1}$ is the difference in those bounds. If $p>0$ then the $\phi_{out}$ is \emph{infeasible}, otherwise it \emph{might be} feasible.

\paragraph{Boolean Abstraction} First, \tool{} creates the boolean variables $v_{00}, v_{01}, v_{10}, v_{11}$ to represent the activation status of the hidden neurons $n_{00}, n_{01}, n_{10}, n_{11}$, respectively.
Next, \tool{} forms the initial clauses $\{ c_0: v_{00} \lor \overline{v_{00}} ; c_1: v_{01} \lor \overline{v_{01}} ; c_2: v_{10} \lor \overline{v_{10}} ; c_3: v_{11} \lor \overline{v_{11}} \}$, which indicate that these variables are either \texttt{true} (active) or \texttt{false} (inactive).

\paragraph{DPLL(T) Iterations} \tool{} now searches for satisfiable activation patterns, i.e., an assignment $\sigma$ over these variables that satisfies the clauses \emph{and} the constraints in the network that they represent. 

\textbf{Iteration 1}: \tool{} starts with an empty assignment, $\sigma_0$. Next \tool{} performs abstraction to obtain the lower and upper bounds $n_{00} \in [-2.1, 0.5]$, $n_{01} \in [-0.6, 1.2]$, $n_{10} \in [-0.42, 0.16]$, and $n_{11} \in [-0.16, 0.9]$. 
Thus $p_{\sigma_0}=-0.1663$ and $\phi_{out}$ might be feasible. 
\tool{} then runs LP solving on these bounds and improves them: $n_{10} \in [-0.3, 0.09]$ and $n_{11} \in [0.0, 0.9]$. 
$n_{11}$ has been determined to be \emph{stable} in its active phase.
The new bound is $p_{\sigma_0}=-0.0942$ and $\phi_{out}$ remains feasible. 

Now \tool{} decides a variable and assigns it a value. 
This decision step performs \emph{neuron splitting}, e.g., the decision $v_i \mapsto T$ means that neuron $n_i$ has a non-negative ReLU value and therefore is active.  Decisions may be inconsistent and require backtracking which is expensive.  \tool{} reduces neuron splitting through neuron stabilization and parallel search.  First, if a neuron, like $n_{11}$, is stable it does not need to be split. Second, \tool{} explores multiple decision branches in parallel.

After determining that $n_{11}$ is stable, we only have to consider the other three neurons. Suppose that \tool{} chooses $v_{01}$. A sequential algorithm  would decide a value for $v_{01}$ and continue with that decision.  Parallel \tool{} mitigates the possibility of wrong decisions by processing both decisions simultaneously. Specifically, \tool{} adds both assignments $\{ \sigma_1 : v_{01}$, $\sigma_2 : \overline{v_{01}} \}$ to the set of assignments to be considered.

\textbf{Iteration 2}:  \tool{} selects both $\sigma_1$ and $\sigma_2$ and runs them in parallel.
As before \tool{} attempts to tighten bounds to determine stable neurons, but none can be found -- our example is too simple. 
In deduction, for the $\sigma_1 : v_{01}$ branch, \tool{} generates a new constraint $n_{01} \in [0.0, 1.2]$, represented by $v_{01} \mapsto T$, corresponding to the active phase of $n_{01} \in [-0.6, 1.2]$. \tool{} then uses the new constraint to compute output bounds $p_{\sigma_1}=-0.0941$ so $\phi_{out}$ is feasible.

Similarly, for the $\sigma_2 : \overline{v_{01}}$ branch, \tool{} creates the constraint $n_{01} \in [-0.6, 0.0]$ and obtains $p_{\sigma_2}=0.0001$ which indicates that $\phi_{out}$ is \emph{not} feasible. This means that the decision $v_{01} \mapsto F$ is inconsistent and in a sequential DPLL(T) we would backtrack to try $v_{01} \mapsto T$. However, in \tool{} we are already processing this decision branch, $\sigma_{1}$, in parallel and therefore do not backtrack. 
Next, \tool{} analyses the infeasible assignment $\sigma_2 : \overline{v_{01}}$ and learns the  new conflict clause $c_4: {v_{01}}$.
                
For the $\sigma_1 : v_{01}$ branch, \tool{} determines feasibility and  chooses $v_{00}$ to split which adds variable assignments $\{ \sigma_3 : v_{01} \land v_{00}$, $\sigma_4 : v_{01} \land \overline{v_{00}} \}$ to the set of assignments to be considered.

\tool{} continues for a few more iterations and determines that all activation patterns are infeasible and returns \texttt{UNSAT}, indicating the desired property is valid.  

While simple, this example illustrates the benefits of stabilization and parallelization relative to a baseline DPLL(T)-based DNN verification approach.
The next section details how both of these techniques work and describes how restarts can improve performance on challenging verification problems.

\section{The \tool{} Approach}\label{sec:alg}

\begin{algorithm}[t]
    \footnotesize
    \DontPrintSemicolon
  
    \Input{DNN $\alpha$, property $\phi_{in} \Rightarrow \phi_{out}$, parallel factors $n$ and $k$}
    \Output{\texttt{unsat} if the property is valid and \texttt{sat} otherwise}
    \BlankLine

    $\clauses \leftarrow \BooleanAbstraction(\alpha)$\; \label{line:boolean_abstraction}

    \While{\true}{\label{line:algostart}
        
        $\assignments \leftarrow [ (\emptyset, \emptyset) ]$ \tcp{initialize empty assignment and \igraph}
        
        \While(\tcp*[h]{main DPLL loop}){\true}{\label{line:dpllstart}
            \tcp{select $n$ assignments (activation patterns) and corresponding $\igraph$s}
            $[(\sigma_1, \igraph_1), ..., (\sigma_n, \igraph_n)] \leftarrow \Select(\assignments, n)$ \label{line:select}
            
            \tcp{process $n$ assignments in parallel}
            \Parfor{$(\sigma_i, \igraph_i)$ \textbf{in} $[(\sigma_1, \igraph_1), ..., (\sigma_n, \igraph_n)]$}{ \label{line:parfor}
                $\isconflict \leftarrow \true$ \;
                \If{$\BCP(\clauses, \sigma_i, \igraph_i)$}{ \label{line:bcp}
                    \If(\tcp*[h]{stabilize with condition}){\StabilizeCondition{}}{\label{line:stabilize_condition}
                    $\Stabilize(\alpha, \phi_{in}, \phi_{out}, \sigma_i, k)$ \tcp{stabilize $k$ neurons} \label{line:stabilize}
                    }
                    \If{\Deduce{$\sigma_i, \alpha, \phi_{in}, \phi_{out}$}}{\label{line:deduce}

                        $(\issat, v_i) \leftarrow \Decide(\alpha, \phi_{in}, \phi_{out}, \sigma_i)$ \tcp{decision heuristic} \label{line:decide}
                        
                        \If{\issat}{
                            \Return{$\sat$} \tcp{consistent and complete assignment}
                        }

                        $\assignments \leftarrow \assignments \cup \{ (\sigma_i \land v_i, \igraph_i) ~;~ (\sigma_i \land \overline{v_i}, \igraph_i) \}$ \;

                        $\isconflict \leftarrow \false$ \tcp{no conflict}
                    }
                    
                    
                }
                
                \If{\isconflict}{
                    
                    $\clauses \leftarrow \clauses \cup \AnalyzeConflict(\igraph_i)$ \tcp{learn conflict clauses} \label{line:analyze_conflict}
                }
                
            }
            
            \If(\tcp*[h]{check \unsat}){$length(\assignments) \equiv 0$}{
                \Return{$\unsat$} \tcp{no more assignment to be processed}
            }
            
            \If(\tcp*[h]{check restart heuristic}){$\Restart{}$}{
                \Break \tcp{restart occurs}
            }

        }\label{line:dpllend}

    }\label{line:algoend}
  \caption{The \tool{} algorithm.}\label{fig:alg}
  
\end{algorithm}

Fig.~\ref{fig:alg} shows the \tool{} algorithm, which takes as input the formula $\alpha$ representing the ReLU-based DNN $N$ and the formulae $\phi_{in}\Rightarrow \phi_{out}$ representing the property to be proved. 
Internally, \tool{} checks the satisfiability of the formula 
\begin{equation}\label{eq:prob}
\alpha \land \phi_{in} \land \overline{\phi_{out}}.
\end{equation}
\tool{} returns \texttt{unsat} if the formula unsatisfiable, indicating  that $\phi$ is a valid property of $N$, and \texttt{sat} if it is satisfiable, indicating the $\phi$ is not a valid property of $N$.

\subsection{DPLL(T)-based DNN Verification}\label{sec:dpll}

\tool{} uses a DPLL(T)-based algorithm to check unsatisfiability. The algorithm consists of  a Boolean abstraction,  standard DPLL components, and a theory solver (T-solver) that is specific to the verification of ReLU DNNs.  




\subsubsection{Boolean Representation} \functiontextformat{BooleanAbstraction}(Fig.~\ref{fig:alg}, line~\ref{line:boolean_abstraction}) encodes the DNN verification problem into a Boolean constraint to be solved.  This step creates Boolean variables to represent the \emph{activation status} of hidden neurons in the DNN.
\tool{} also forms a set of initial clauses ensuring that each status variable is either \texttt{T} (active) or \texttt{F} (inactive).

\subsubsection{DPLL search} \tool{} iteratively searches for an assignment satisfying the clauses. 
Throughout it maintains several state variables including: $\clauses$, a set of \emph{clauses} consisting of the initial activation clauses and learned conflict clauses;   $\sigma$, a \emph{truth assignment} mapping status variables to truth values which encodes a partial activation pattern; and $igraph$, an \emph{implication graph} used for analyzing conflicts.

\functiontextformat{Decide} (Fig.~\ref{fig:alg}, line~\ref{line:decide}) chooses an unassigned variable and assigns it a random truth value.
Assignments from \functiontextformat{Decide} are essentially guesses that can be wrong which degrades performance. 
The purpose of \functiontextformat{BCP}, \functiontextformat{Deduce}, and \functiontextformat{Stabilize} -- which are discussed below -- is to eliminate unassigned variables so that \functiontextformat{Decide} has fewer choices.

\functiontextformat{BooleanConstraintPropagation} or \functiontextformat{BCP} (Fig.~\ref{fig:alg}, line~\ref{line:bcp}) detects \emph{unit clauses}\footnote{A unit clause is a clause that has a single unassigned literal.} from constraints representing the current assignment and clauses and infers values for variables in these clauses.
For example, after the decision $a\mapsto F$, \functiontextformat{BCP} determines that the clause $a\vee b$ becomes unit, and infers that $b \mapsto T$.
Internally, \tool{} uses an \emph{implication graph}~\cite{barrett2013decision} to represent the current assignment and the reason for each \functiontextformat{BCP} implication. 

\functiontextformat{AnalyzeConflict}  (Fig.~\ref{fig:alg}, line~\ref{line:analyze_conflict}) processes an implication graph with a conflict to learn a new \emph{clause} that explains the conflict.
The algorithm traverses the implication graph backward, starting from the conflicting node, while constructing a new clause through a series of resolution steps.
\functiontextformat{AnalyzeConflict} aims to obtain an \emph{asserting} clause, which is a clause that will result a \functiontextformat{BCP} implication.
These are added to $\clauses$ so that they can block further searches from encountering an instance of the conflict.

These are standard components in DPLL-based algorithms including modern SAT/SMT solvers and \neuralsat{}. As shown in Fig.~\ref{fig:neuralsat}, DPLL also has backtracking, which allows the algorithm to go back to an incorrect assignment decision and make the correct one. However, as will be described in \S\ref{sec:parallel}, the \tool{} parallel DPLL(T) does not require backtracking because it has optimistically considered both the correct and incorrect assignments simultaneously. 

\subsubsection{Theory Solver}\label{sec:theory_solver}
\tool{}'s Theory or T-solver (Fig.~\ref{fig:alg}, lines 9-16) consists of two parts: stabilization and deduction.

\functiontextformat{Deduce} (Fig.~\ref{fig:alg}, line~\ref{line:deduce}) checks the feasibility of the DNN constraints represented by the current propositional variable assignment. 
This component is shared with \neuralsat{} and it leverages specific information from the DNN problem, including input and output properties, for efficient feasibility checking.  
Specifically, it obtains neuron bounds using the polytope abstraction\cite{xu2020fast} and performs infeasibility checking to detect conflicts.

The second part of the theory solver, which is specific to \tool{}, implements stabilization and is described next.

\subsection{Improvements in \tool{}}

We now describe neuron stability, parallel search, and restart. In \S\ref{sec:rq1} and \S\ref{sec:rq2} we present ablation studies demonstrating the performance of these ideas individually and in combination.

\subsubsection{Neuron Stability}\label{sec:stabilization} 
The key idea in using neuron stability is that if we can determine that a neuron is stable,  we can assign the exact truth value for the corresponding Boolean variable instead of having to guess. This has a similar effect as \functiontextformat{BCP} -- reducing mistaken assignments by \functiontextformat{Decide} -- but it operates at the theory level not the propositional Boolean level.

Stabilization involves the solution of a mixed integer linear program (MILP) system~\cite{tjeng2019evaluating}:
\begin{equation}
    \begin{aligned}
        &\mbox{(a)}\quad z^{(i)} = W^{(i)} \hat{z}^{(i-i)} + b^{(i)}; \\
        &\mbox{(b)}\quad y = z^{(L)};  x = \hat{z}^{(0)}; \\
        &\mbox{(c)}\quad \hat{z}_j^{(i)} \ge {z}_j^{(i)}; \hat{z}_j^{(i)} \ge 0; \\
        &\mbox{(d)}\quad a_j^{(i)} \in \{ 0, 1\} ;\\
        &\mbox{(e)}\quad \hat{z}_j^{(i)} \le {a}_j^{(i)} {u}_j^{(i)}; \hat{z}_j^{(i)} \le {z}_j^{(i)} - {l}_j^{(i)}(1 - {a}_j^{(i)}); \\
    \end{aligned}
    \label{eq:mip}
\end{equation}
where $x$ is input, $y$ is output, and $z^{(i)}$, $\hat{z}^{(i)}$, $W^{(i)}$, and $b^{(i)}$ are the pre-activation, post-activation, weight, and bias vectors for layer $i$. 
The equations encode the semantics of a DNN as follows:
(a) defines the affine transformation computing the pre-activation value for a neuron in terms of outputs in the preceding layer;
(b) defines the inputs and outputs in terms of the adjacent hidden layers;
(c) asserts that post-activation values are non-negative and no less than pre-activation values;
(d) defines that the neuron activation status indicator variables that are either 0 or 1; and
(e) defines constraints on the upper, $u_j^{(i)}$, and lower, $l_j^{(i)}$, bounds of the pre-activation value of the $j$th neuron in the $i$th layer.
Deactivating a neuron, $a_j^{(i)} = 0$, simplifies the first of the (e) constraints to $\hat{z}_j^{(i)} \le 0$, and activating a neuron simplifies the second to $\hat{z}_j^{(i)} \le z_j^{(i)}$, which is consistent with the semantics of $\hat{z}_j^{(i)} = max(z_j^{(i)},0)$.


Fig.~\ref{fig:stabilize} describes $\Stabilize$ solves this equation system.
First,  a MILP problem is created from the current assignment, the DNN, and the property of interest using formulation in Eq.~\ref{eq:mip}.
Note that the neuron lower (${l}_j^{(i)}$) and upper bounds (${u}_j^{(i)}$) can be quickly computed by polytope abstraction.

Next, it collects a list of all unassigned variables which are candidates being stabilized (line~\ref{line:find}). 
In general, there are too many unassigned neurons, so
\functiontextformat{Stabilize} restricts consideration to $k$ candidates.
Because each neuron has a different impact on abstraction precision we prioritize the candidates. 
In \functiontextformat{Stabilize}, neurons are prioritized based on their interval boundaries (line~\ref{line:sort}) with a preference for neurons with either lower or upper bounds that are closer to zero.
The intuition is that neurons with bounds close to zero are more likely to become stable after tightening.

We then select the top-$k$ (line~\ref{line:topk}) candidates and seek to further tighten their interval bounds.
The order of optimizing bounds of select neurons is decided by its boundaries, e.g., if the lower bound is closer to zero than the upper bound then the lower bound would be optimized first.
These optimization processes, i.e., \functiontextformat{Maximize} (line~\ref{line:maximize1} or line~\ref{line:maximize2}) and \functiontextformat{Minimize} (line~\ref{line:minimize1} or line~\ref{line:minimize2}), are performed by an external LP solver (e.g., Gurobi~\cite{gurobi}).

Note that the work in~\cite{tjeng2019evaluating} uses the MILP system in Eq.~\ref{eq:mip} to encode the entire verification problem and thus is limited to the encodings of small networks that can be handled by an LP solver.
In contrast, \tool{} creates this system based on the current assignment, which has significantly fewer constraints.
Moreover, we only use the computed bounds of hidden neurons from this system, and thus even if it cannot be solved, \tool{} will still continue.

\begin{algorithm}[t]
    \footnotesize

    \Input{DNN $\alpha$, property $\phi_{in} \Rightarrow \phi_{out}$, current assignment $\sigma$, number of neurons for stabilization $k$}
    \Output{Tighten bounds for variables \textbf{not} in $\sigma$ (unassigned variables)}
    \BlankLine

    $\model \leftarrow \MIP(\alpha, \phi_{in}, \phi_{out}, \sigma)$ \tcp{create model (Eq.~\ref{eq:mip}) with  current assignment}

    $[v_1, ..., v_m] \leftarrow \GetUnassignedVariable(\sigma)$ \tcp{get all $m$ current unassigned variables} \label{line:find}

    $[v_1', ..., v_m'] \leftarrow \Sort([v_1, ..., v_m])$ \tcp{prioritize tightening order} \label{line:sort}
    $[v_1', ..., v_k'] \leftarrow \Select([v_1', ..., v_m'], k)$ \tcp{select top-$k$ unassigned variables, $k \le m$} \label{line:topk}

    \tcp{stabilize $k$ neurons in parallel}
    \Parfor{$v_i$ \textbf{in} $[v_1', ..., v_k']$}{
        \If(\tcp*[h]{lower is closer to 0 than upper, optimize lower first}){$(v_i.lower + v_i.upper) \ge 0$}{
            $\Maximize(\model, v_i.lower)$ \tcp{tighten lower bound of $v_i$} \label{line:maximize1}
            \If(\tcp*[h]{still unstable}){$v_i.lower < 0$}{
                $\Minimize(\model, v_i.upper)$ \tcp{tighten upper bound of $v_i$} \label{line:minimize1}
            }
        }
        \Else(\tcp*[h]{upper is closer to 0 than lower, optimize upper first}){
            $\Minimize(\model, v_i.upper)$ \tcp{tighten upper bound of $v_i$} \label{line:minimize2}
            \If(\tcp*[h]{still unstable}){$v_i.upper > 0$}{
                $\Maximize(\model, v_i.lower)$ \tcp{tighten lower bound of $v_i$} \label{line:maximize2}
            }
        }
    }
    \vspace*{-0.2in}
    \caption{\texttt{Stabilize}}
    \label{fig:stabilize}
\end{algorithm}

\subsubsection{Parallelism}\label{sec:parallel}
The DPLL(T) process in \tool{} is designed as a tree-search problem where each internal node encodes an \textit{activation pattern} defined by the variable assignments from the root.
To parallelize DPLL(T), we adopt a beam search-like strategy which combines distributed search from Distributed Tree Search (DTS) algorithm~\cite{ferguson1988distributed} and Divide and Conquer (DNC)~\cite{le2017painless} paradigms for splitting the search space into disjoint subspaces that can be solved independently.
At every step of the search algorithm, we select up to $n$ nodes  of the DPLL(T) search tree to create a beam of width $n$.
This splits (like DNC) the search into $n$ subproblems that are independently processed.  Each subproblem extends the tree by a depth of 1.

Our approach simplifies the more general DNC scheme since the $n$ bodies of the \textbf{parfor} on line~\ref{line:parfor} of Fig.~\ref{fig:alg} are roughly load balanced.
While this is a limited form of parallelism, it sidesteps one of the major roadblocks to DPLL parallelism -- the need to efficiently synchronize across load-imbalanced subproblems~\cite{le2017painless,le2019modular}.

In addition to raw speedup due to multiprocessing, parallelism accelerates the sharing of information across search subspaces, in particular learned clause information for DPLL. In \tool{}, we only generate independent subproblems which eliminates the need to coordinate their solution. When all subproblems are complete, their conflicts are accumulated, Fig.~\ref{fig:alg} line~\ref{line:analyze_conflict},  to inform the next round of search.
As we show in \S\ref{sec:eval}, the engineering of this form of parallelism in DPLL(T) leads to substantial performance improvement.

\subsubsection{Restart}

\SetKwFunction{threshold}{threshold}
\SetKwFunction{maxvisited}{max\_visited}
\SetKwFunction{maxremaining}{max\_remaining}
\SetKwData{implicationgraph}{igraph}

As with any stochastic algorithm, \tool{} would perform poorly if it gets into a subspace of the search that does not quickly lead to a solution, e.g., due to choosing a bad sequence of neurons to split~\cite{ferrari2022complete,wang2021beta,de2021improved} .  
This problem, which has been recognized in early SAT solving, motivates the introduction of restarting the search~\cite{gomes1998boosting} to avoid being stuck in such a \emph{local optima}. 


\tool{} uses a simple restart heuristic that triggers a restart when either the number of processed assignments (nodes) exceeds a pre-defined number or the number of remaining assignments that need be checked exceeds a pre-defined threshold. 
After a restart, \tool{} avoids using the same decision order of previous runs (i.e., it would use a different sequence of neuron splittings). It also resets all internal information except the learned conflict clauses, which are kept and reused as these are \textit{facts} about the given constraint system.
This allows a restarted search to quickly prune parts of the  space of assignments.
Although restarting may seem like an engineering aspect, it plays a crucial role in stochastic algorithms like \tool{} and helps reduce verification time for challenging problems as shown in \S\ref{sec:rq1}.



\subsection{\tool{} Implementation}
\tool{} is written in Python, and uses Gurobi~\cite{gurobi} for LP solving and bounds tightening, and the LiRPA abstraction library~\cite{xu2020fast} for approximation.
Currently, \tool{} supports feedforward (FNN), convolutional (CNN), and Residual Learning Architecture (ResNet) neural networks that use ReLU.
\tool{} supports the standard specification formats ONNX~\cite{onnx2} for neural networks and VNN-LIB~\cite{vnnlib} for properties.
These formats are standard and are supported by state-of-the-art DNN verification tools, which enable comparative evaluation.

\section{Experimental Design}
Our goals are to understand how incorporating stabilization and other DPLL(T) optimizations allows for scaling of DNN verification.
We focus our evaluation on the following research questions:
\noindent\mbox{~~~~}\textbf{RQ1} (\S\ref{sec:rq1}): How does stabilization impact the performance of DPLL(T)-based DNN verification?  
\noindent\mbox{~~~~}\textbf{RQ2} (\S\ref{sec:rq2}): How do \tool{} optimizations improve performance in isolation and combination?
\noindent\mbox{~~~~}\textbf{RQ3} (\S\ref{sec:rq3}): How does \tool{} compare to state-of-the-art DNN verifiers?

\begin{table*}
    \footnotesize
    \caption{Benchmark instances. U: \texttt{unsat}, S: \texttt{sat}, ?: \texttt{unknown}.}\label{tab:benchmarks}
    \vspace*{-3mm}
    
    \begin{tabular}{c|cccc|cc}
              \toprule
            \multirow{2}{*}{\textbf{Benchmarks}} &\multicolumn{2}{c}{\textbf{Networks}} &  \multicolumn{2}{c|}{\textbf{Per Network}} &\multicolumn{2}{c}{\textbf{Tasks}}\\
            & Type & Networks & Neurons & Parameters & Properties & Instances (U/S/?)\\
  
              \midrule
              ACAS Xu                       & FNN          & 45               & 300   & 13305     & 10 & 139/47/0 \\
              \midrule
              
              MNISTFC                       & FNN          & 3                &  0.5-1.5K    & 269-532K    & 90 & 56/23/11\\
              \midrule
              CIFAR2020                     & CNN          & 3                &  17-62K    & 2.1-2.5M    & 203 & 149/43/11 \\
              \midrule
              
              RESNET\_A/B                     & CNN+ResNet   & 2           & 11K & 354K    & 144 & 49/23/72 \\
              \midrule
              MNIST\_GDVB                   & FNN          & 38               & 0.7-5.1K & 0.2-3.0M & 16 & 51/0/39 \\
                                                
            \midrule
      \textbf{Total}       &                & \textbf{91}  &   &                   & \textbf{463}      & \textbf{444/136}/133 \\
              \bottomrule
          \end{tabular}
  \end{table*}

\subsection{Benchmarks: DNN Verification Problems}
To gain insights into the performance improvements of \tool{} we require benchmarks that force the algorithm to search a non-trivial portion of the space of activation patterns.   It is well-known that SAT problems can be very easy to solve regardless of their size or whether they are satisfiable or unsatisfiable~\cite{gent1994sat}.  The same is true for DNN verification problems.
The organizers of the first three DNN verifier competitions remark on the need for benchmarks that are ``not so easy that every tool can solve all of them'' in order to assess verifier performance~\cite{brix2023first}.  

To achieve this we leverage a systematic DNN verification problem
generator GDVB~\cite{xu2020systematic}.  GDVB takes a seed neural network as input and systematically varies a number of architectural parameters, e.g., number of layers, and neurons per layer, to produce a set of DNNs.  In this experiment, we
begin with a single MNIST network with 3 layers, each with 1024 neurons
and generate 38 different DNNs that cover combinations of parameter variations.   We leverage the fact that local robustness properties are a pseudo-canonical form for pre-post condition specifications~\cite{shriver2021reducing} and use GDVB
to generate 16 properties with varying radii and center points.
Next we run two state-of-the-art verifiers: \crown{} and \mnbab{}, for each of the $38*16 = 608$ combinations of DNN and property with a small timeout of 200 seconds.  Any problem that could be solved within that timeout was removed from the benchmark as ``too easy''.
This resulted in 90 verification problems that not only are more computationally challenging than benchmarks used in other studies, e.g.,~\cite{muller2022third}, but also exhibit significant architectural diversity.
We use this \textbf{MNIST\_GDVB} benchmark for RQ1 and RQ2 to study the variation in performance on challenging problems.

For RQ3 we use five \vnncomp{} standard benchmarks in addition to MNIST\_GDVB.
These benchmarks, shown in Tab.~\ref{tab:benchmarks}, consist of 91 networks, spanning multiple types and architectures of layers, and 463 safety and robustness properties.  
The \textbf{Per Network} column gives the size of each network (\textbf{neurons} are the numbers of hidden neurons and \textbf{parameters} are the numbers of weights and biases). For example, each FNN in ACAS Xu has 5 inputs, 6 hidden layers (each with 50 neurons), 5 outputs, and thus has 300 neurons ($6 \times 50$) and 13305 parameters ($5\times 50\times 50 + 2\times 50 \times 5 + 6 \times 50 + 5$). 

In total, we have 713 problem instances (an instance is the verification task of a property of a network).
Among these instances, 444 are known to be \texttt{unsat} (U), 136 are \texttt{sat} (S), and 133 are unknown (?) because no existing verifiers, in this study or in VNN-COMP, can solve them. We exclude unknown instances from our study because they do not contribute to our evaluation or comparison to other tools.

The six benchmarks are as follows. 
\textbf{ACAS Xu} consists of 45 FNNs to issue turn advisories to aircrafts to avoid collisions. Each FNN has 5 inputs (speed, distance, etc). 
We use all 10 safety properties as specified in~\cite{katz2017reluplex} and \vnncomp{}, where properties 1--4 are used on 45 networks and properties 5--10 are used on a single network.
\textbf{MNISTFC} consists of 3 FNNs for handwritten digit recognition and 30 robustness properties.
Each FNN has 28x28 inputs representing a handwritten image.
\textbf{CIFAR2020} has 3 CNNs for object detection and 203 robustness properties (each CNN has a set of different properties). Each network uses 3x32x32 RGB input images.
For \textbf{RESNET\_A/B}, each benchmark has only one network with the same architecture and 72 robustness properties. Each network uses 3x32x32 RGB input images. 

\subsection{Baselines: DNN Verifiers} 
For RQ1 we compare \tool{} to 
\textbf{\neuralsat{}}~\cite{duong2023dpllt} which is the only DPLL(T) DNN verifier available\footnote{\url{https://github.com/dynaroars/neuralsat-solver}}. 
\neuralsat{} is recent and did not participate in \vnncomp{}. However, it has been shown to have good performance for feedforward networks.

RQ2 compares different configurations of \tool{} to each other.

For RQ3, we selected four well-known DNN verifiers as baselines for comparison in addition to \neuralsat{}.
\textbf{\crown{}}~\cite{wang2021beta,zhang2022general} employs multiple abstractions and algorithms for efficient analysis, e.g., input splitting for networks with small input dimensions and parallel Branch-and-Bound~\cite{bunel2020branch} (BaB) otherwise. 
\textbf{\mnbab{}}\cite{ferrari2022complete}, the successor of \eran{}\cite{singh2019abstract,singh2019beyond}, uses multiple abstractions and BaB.
\textbf{\marabou{}}~\cite{katz2019marabou,katz2022reluplex}, the successor of the Reluplex work,  
is a simplex-based solver that employs a parallel Split-and-Conquer (SnC)~\cite{wu2020parallelization} search  and uses polytope abstraction~\cite{singh2019abstract} and LP-based bound tightening.
\textbf{\nnenum{}}~\cite{bak2020improved} combines optimizations such as parallel case splitting and multiple levels of abstractions, e.g., 
three types of zonotopes with imagestar/starset~\cite{tran2019star}.

These four tools competed in \vnncomp{}~\cite{muller2022third} and were among the very top performers.
For example, 
\crown{} is the winner for MNISTFC and also the overall winner, 
\mnbab{} ranked 3rd on MNISTFC and second overall,
and \nnenum{} was the only one that can solve all instances in ACAS Xu and was 4th overall.  
\marabou{}  ranked 6th on MNISTFC and 7th overall. 

\subsection{Experimental Setup}\label{sec:expsettings}
Our experiments were run on a Linux machine with an AMD Threadripper 64-core 4.2GHZ CPU, 128GB RAM, and an NVIDIA GeForce RTX 4090 GPU with 24 GB VRAM. 
All tools use multiprocessing (even external tools/libraries including Gurobi, LiRPA, and Pytorch are multi-thread).  
\crown{}, \mnbab{}, \neuralsat{}, and \tool{} leverage GPU processing for abstraction.

To conduct a fair evaluation, 
we reuse the benchmarks and
installation/run-scripts available from VNN-COMP\footnote{\url{https://github.com/ChristopherBrix/vnncomp2022_benchmarks}}. 
These scripts were tailored by the developers of each verifier to maximize performance on each benchmark.
VNN-COMP uses varying runtimes for each problem instance ranging from 30 seconds to more than 20 minutes. The competition also uses several different Amazon AWS instances with different configurations (e.g., CPU, GPU, RAM) to run the tools.
Thus, we experimented with timeouts on our machine and settled on 900 seconds per instance which allowed the verifiers to achieve similar scoring performance reported in VNN-COMP'22.

\section{Results and Analysis}\label{sec:eval}
We discuss the metrics for each question, present experimental results, and interpret those results to answer the research questions.

\subsection{RQ1: Benefit of Stabilization}\label{sec:rq1}

We focus here on the benefit of stabilization on DPLL(T)-based DNN verification as implemented in \neuralsat{}.  We use the 51 challenging verification problems in the MNIST\_GDVB benchmark to explore performance and measure the number of problems solved and the time to solve problems as metrics.
\begin{figure}[t]
    \hfill
    \begin{subfigure}[b]{0.47\textwidth}
        \centering
        \footnotesize
        \begin{tabular}{c|c|c|c}
            \toprule
            \textbf{Tool} & \textbf{Setting} & \textbf{\#Solved} &\textbf{Avg. Time} \\
            \midrule
            \neuralsat{}             & -       & 4    & 867.35 \\
            \midrule
            \multirow{5}{*}{\tool{}} & S       & 6    & 833.25 \\
                                     & P       & 14   & 713.21 \\
                                     & P+R     & 17   & 741.00 \\
                                     & P+S     & 38   & 430.60 \\
                                     & P+S+R   & 48   & 330.46 \\
            \bottomrule    
       \end{tabular}
       \vspace*{4mm}
        \caption{Problems solved and solve time (s).}\label{tab:ablation}
     \end{subfigure}
     \hfill
     \begin{subfigure}[b]{0.47\textwidth}
         \centering
        \includegraphics[width=\linewidth]{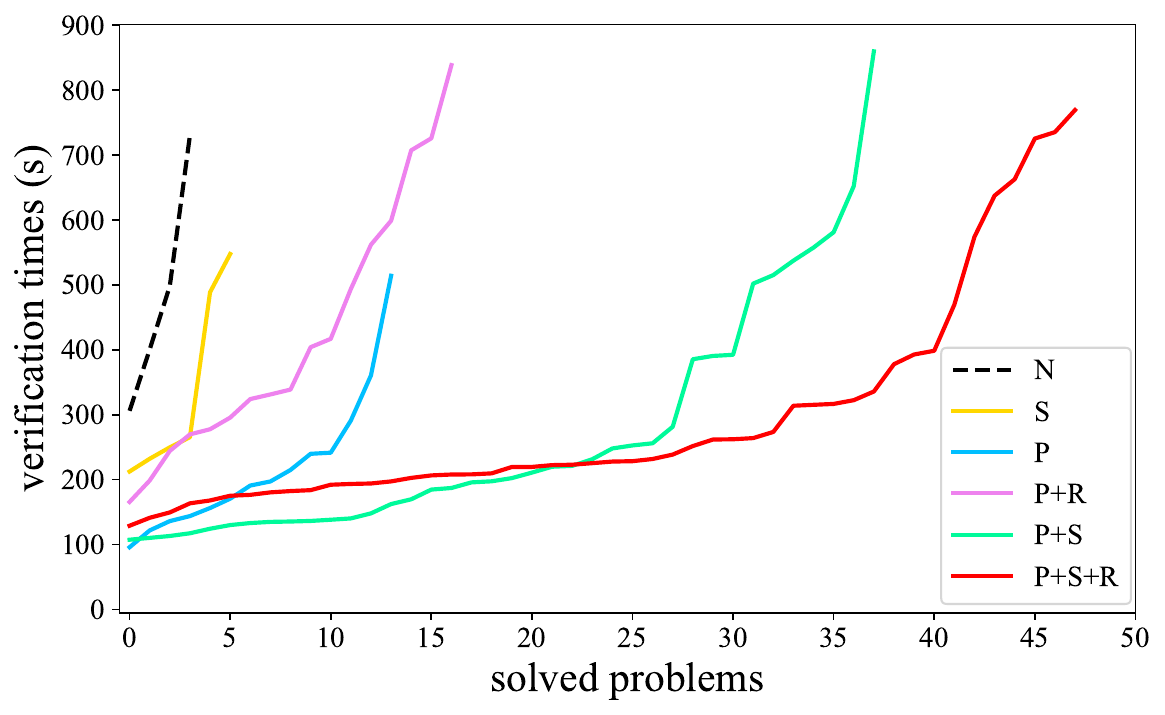}
        \caption{Sorted solved problems}\label{fig:ablation}
     \end{subfigure}
     \hfill
     \vspace{-2mm}
    \caption{Performance of \tool{} with different optimization settings in comparison to \neuralsat{} on the \textbf{MNIST\_GDVB} benchmark with 900 second timeout, where:  ``N'' is the base case (\neuralsat{}), ``P'' enables Parallelism, ``R'' enables Restarts, and ``S'' enables Stabilization.}\label{fig:rq1rq2}
\end{figure}

Fig.~\ref{fig:rq1rq2} presents data on \neuralsat{}: the first row in the table on the left and the black dashed line in the plot on the right.  The plot on the right shows the problems solved
within the 900-second timeout for each technique sorted by runtime from fastest to slowest; problems that timeout are not shown on the plot.
Enabling only stabilization in \tool{} yields the data indicated with an ``S'': the second row and yellow lines, respectively.
We observe a 50\% increase in the number of problems solved with stabilization.  The average times show a modest reduction of about 4\%, but since \neuralsat{} or ``S'' solved just a few benchmarks the average is swamped by the time taken by problems that timeout -- at 900 seconds.  Comparing the dashed and yellow lines in Fig.~\ref{fig:ablation} shows that for the solved problems ``S'' reduces verification time significantly, e.g., on the first problem from just over 300 seconds to just over 200 seconds.
Stabilization alone improves performance, but it has a much more significant benefit in combination with other optimizations.   

\begin{figure}
\hfill
\begin{subfigure}[b]{0.5\textwidth}
\centering
    \includegraphics[width=0.9\linewidth]{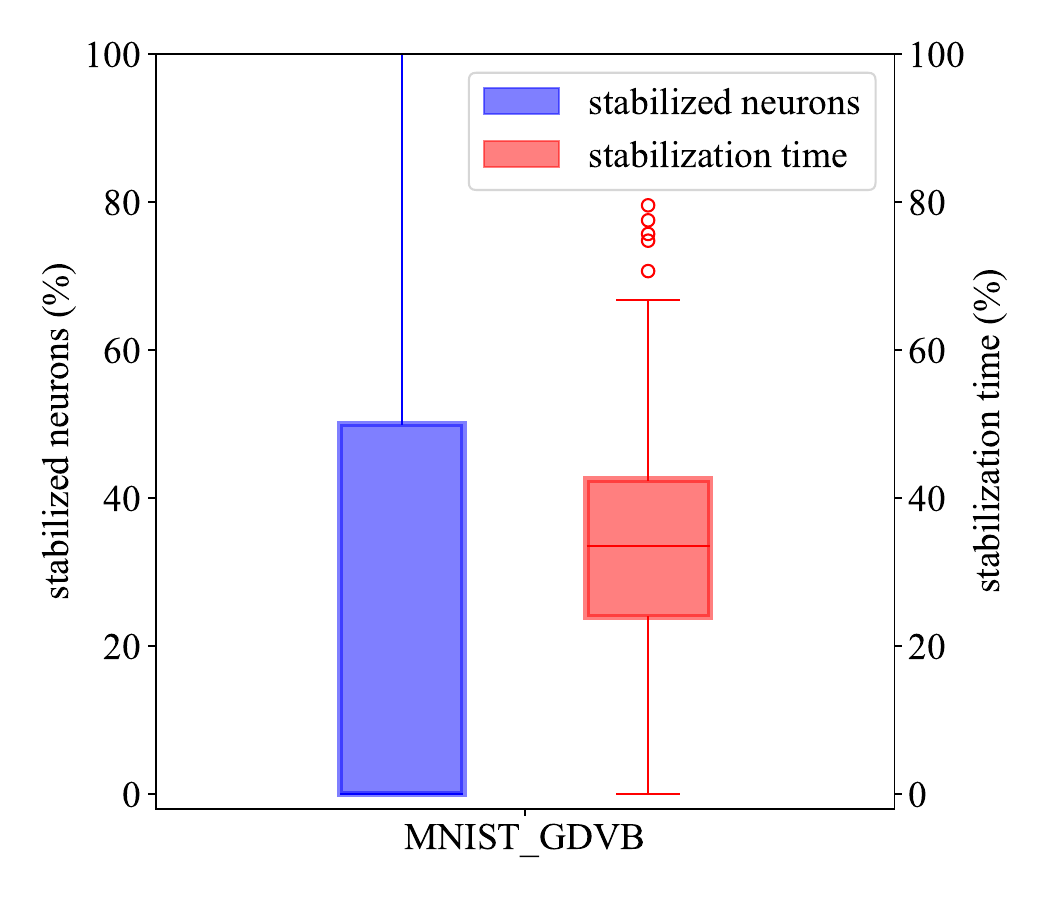}
    \vspace{-2mm}
    \caption{Stabilization rate (per call) and stabilization time}\label{fig:ratetime}
\end{subfigure}
\hfill
\begin{subfigure}[b]{0.48\textwidth}
\centering
    \includegraphics[width=0.8\linewidth]{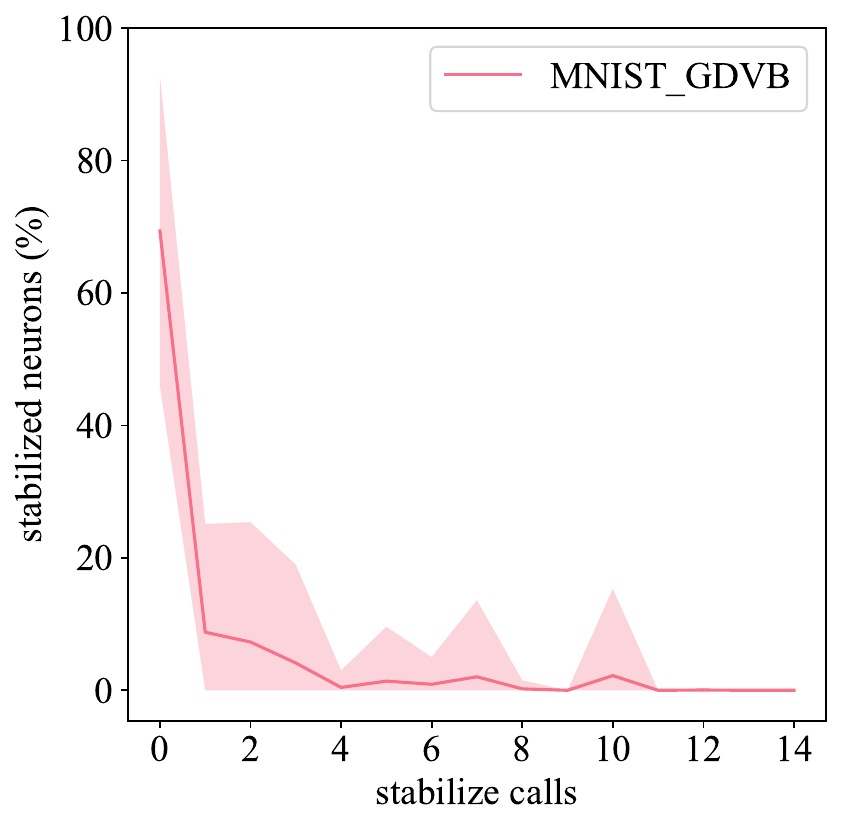}
    \vspace{-2mm}
    \caption{Stabilization rate over time}\label{fig:rateovertime}
\end{subfigure}
\hfill
\vspace{-2mm}
\caption{Stabilization cost and effectiveness during verification.}
\end{figure}

We collected data to understand how frequently neurons could be stabilized and at what cost.
Fig.~\ref{fig:ratetime} plots the percentage of neurons that are stabilized across the
MNIST\_GDVB benchmark, on the left axis, and the percentage of verification time taken up
by stabilization, on the right axis.  This aggregated data shows that stabilization can incur a non-trivial share of verification time, but as the data in Fig.~\ref{fig:rq1rq2} showed despite this overhead the overall verification time is reduced for solved problems.  

We can also observe that while the mean number of stabilized neurons is low, the variance is quite high which indicates a degree of effectiveness in reducing the combinatorics in subsequent searches.
We dug into the stabilization data further to try to understand this variance.
Fig.~\ref{fig:rateovertime} plots the mean -- red line -- and standard deviation -- shaded region -- of the number of stabilized neurons over time during verification; recall from line~\ref{line:stabilize_condition} of Fig.~\ref{fig:alg} that stabilization is selectively enabled during search.
Stabilization is effective early in the search and less so as it progresses.
This makes sense since line \ref{line:sort} in Fig.~\ref{fig:stabilize} prioritizes neurons for stabilization.
This is desirable because it encourages stabilization at the beginning of the search which leads to a greater combinatorial reduction in the search and a consequent improvement in its scalability.

\begin{tcolorbox}[left=1pt,right=1pt,top=1pt,bottom=1pt]
\textbf{RQ1 Findings}: Stabilization improves the number of problems solved and reduces verification time. It does so by trading overhead to compute stable neurons to linearize parts of the search of the space of activation patterns.  Moreover, it pushes this linearization to the top of the search tree to yield greater combinatorial reduction.
\end{tcolorbox}

\subsection{RQ2: Optimization Ablation Study}\label{sec:rq2}
We used the same benchmark as in RQ1, but here we focus primarily on the benefits and interactions among the optimizations in \tool{}. 
The bottom five rows in the table on the left of Fig.~\ref{fig:rq1rq2} 

We omit the use of restart on its own, since it is intended to function in concert with parallelization.
Both ``S'' and ``P'' improve the number of problems solved and reduce cost relative to the \neuralsat{} baseline, but parallelism yields greater improvements.  
When parallelism is combined with restart we see that the number of problems solved increases, but the average time increases slightly.
The reason for this is that for the 3 additional benchmarks that could be solved the verification process had conducted a partial search of the space of activation patterns prior to restarts and the cost of that search is added to the cost of the successful post-restart search.

Perhaps most noteworthy is the data on parallelism in combination with stabilization.
We see a significant jump in the number of solved problems relative to both ``S'' and ``P'' -- a 6.3 fold and 2.7 fold increase, respectively.  As illustrated in Fig.~\ref{fig:tree} this combination is synergistic because stabilization creates a \textit{narrower} tree within which the parallel \textit{beam} can make more rapid progress.  Adding in restart yields the best performance in terms of both problems solved -- 12 fold increase \neuralsat{} -- and solve time -- 2.6 fold decrease.

The plot on the right of Fig.~\ref{fig:rq1rq2} shows the trend in verification solve times for each optimization combination across the benchmarks.
One can observe that adding more optimizations improves performance both by the fact that the plots are lower and extend further to the right.
For example, extending ``P'' to ``P+S'' shows lower solve times for the first 17 problems -- the one's ``P'' could solve -- and that 38 of the 51 benchmark problems are solved.
Extending ``P+S'' to the full set of optimizations exhibits what appears to be a degradation in performance for the first 23 problems solved and this is likely due to the fact that, as explained above, restart forces some re-exploration of the search.   However, the benefit of restart shows in the ability to significantly reduce 
verification time for 25 of the 48 problems solved by ``P+S+R''.

\begin{tcolorbox}[left=1pt,right=1pt,top=1pt,bottom=1pt]
\textbf{RQ2 Findings}: Each of the \tool{} optimizations improves on the performance of the baseline DPLL(T)-based DNN verifier.  Moreover, combinations of the optimizations appear to operate synergistically to increase performance beyond their additive benefits.  When running \tool{}, enabling all optimizations appears to be the best choice.
\end{tcolorbox}

\subsection{RQ3: Comparison with state-of-the-art DNN verifiers}\label{sec:rq3}
In this section, we evaluate \tool{} relative to a set of 5 baseline DNN verifiers across a broader benchmark that reflects the problems used in VNN-COMP~\cite{muller2022third}.
For metrics, we adopt the scoring system proposed for VNN-COMP 2023 which seeks to balance the relative difficulty
of verifying a problem versus falsifying it and to account for the possibility that verifiers report erroneous results.   More specifically, for each benchmark instance, a verifier scores 10 points if it correctly verifies an instance, 1 point if it correctly falsifies an instance, 
    0 points if it cannot solve (e.g., times out, has errors, or returns \texttt{unknown}), and -150 points if it gives incorrect results\footnote{We note that all of the verifiers in our study gave correct results on the considered benchmarks.}. This scoring emphasizes a technique's ability to correctly verify problems\footnote{We dropped the extra 2 bonus points for the fastest verifiers in the VNN-COMP'22 scoring system because VNN-COMP has removed this time bonus as they found it did not make a difference in scoring}.

\begin{table}[]
     \caption{A \textbf{Verifier}'s rank (\textbf{\#}) is based on its VNN-COMP score (\textbf{S}) on a {benchmark}. For each benchmark, the number of problems verified (\textbf{V}) and falsified (\textbf{F}) are shown.}\label{tab:score}

\resizebox{\textwidth}{!}{
\begin{tabular}{c|cccc|cccc|cccc|cccc|cccc|cccc}
\toprule
\multirow{2}{*}{\textbf{Verifier}} &
  \multicolumn{4}{c|}{\textbf{ACAS Xu}} &
  \multicolumn{4}{c|}{\textbf{MNISTFC}} &
  \multicolumn{4}{c|}{\textbf{CIFAR2020}} &
  \multicolumn{4}{c|}{\textbf{RESNET\_A/B}} &
  \multicolumn{4}{c|}{\textbf{MNIST\_GDVB}} &
  \multicolumn{4}{c}{\textbf{Overall}} \\

 & \textbf{\#} & \textbf{S} & \textbf{V} & \textbf{F} &
   \textbf{\#} & \textbf{S} & \textbf{V} & \textbf{F} &
   \textbf{\#} & \textbf{S} & \textbf{V} & \textbf{F} &
   \textbf{\#} & \textbf{S} & \textbf{V} & \textbf{F} &
   \textbf{\#} & \textbf{S} & \textbf{V} & \textbf{F} &
   \textbf{\#} & \textbf{S} & \textbf{V} & \textbf{F} \\
\midrule

\tool{} &
\textbf{1} & \textbf{1437} & \textbf{139} & \textbf{47} & 
2 & 573 & 55 & \textbf{23} & 
\textbf{1} & \textbf{1533} & \textbf{149} & \textbf{43} & 
\textbf{1} & \textbf{513} & \textbf{49} & \textbf{23} & 
\textbf{1} & \textbf{480} & \textbf{48} & 0 & 
\textbf{1} & \textbf{4536} & \textbf{440} & \textbf{136} \\  
\midrule

\crown{} &
3 & 1436 & \textbf{139} & 46 & 
\textbf{1} & \textbf{582} & \textbf{56} & 22 & 
2 & 1522 & 148 & 42 & 
\textbf{1} & \textbf{513} & \textbf{49} & \textbf{23} & 
2 & 400 & 40 & 0 & 
2 & 4453 & 432 & 133 \\ 

\midrule

\neuralsat{} &
5 & 1417 & 137 & \textbf{47} & 
4 & 383 & 36 & \textbf{23} & 
4 & 1522 & 148 & 42 & 
3 & 483 & 46 & \textbf{23} & 
4 & 40 & 4 & 0 & 
3 & 3845 & 371 & 135 \\ 

\midrule

\mnbab{} &
6 & 1097 & 105 & \textbf{47} & 
5 & 370 & 36 & 10 & 
3 & 1486 & 145 & 36 & 
4 & 363 & 34 & \textbf{23} & 
3 & 200 & 20 & 0 & 
4 & 3516 & 340 & 116 \\ 

\midrule

\nnenum{} &
\textbf{1} & \textbf{1437} & \textbf{139} & \textbf{47} & 
3 & 403 & 39 & 13 & 
5 & 518 & 50 & 18 & 
- & - & - & - & 
- & - & - & - & 
5 & 2358 & 228 & 78 \\ 
\midrule

\marabou{} &
4 & 1426 & 138 & 46 & 
6 & 370 & 35 & 20 & 
- & - & - & - & 
- & - & - & - & 
- & - & - & - & 
6 & 1796 & 173 & 66 \\ 

\bottomrule

\end{tabular}
}
\end{table}

Tab.~\ref{tab:score} shows the results of all six tools.  
Since the magnitude of the score is not easily interpreted, since it depends on the size of the benchmark, we report
the \textbf{Rank} of each tool using the VNN-COMP score for each benchmark as well as the overall rank.
Tools that do not work on a benchmark are not shown under that benchmark (e.g., \marabou{} reports errors for all CIFAR2020 problems, \nnenum{} and \marabou{} cannot solve any instances of MNIST\_GDVB).
The last two columns break down the number of problems each verifier was able to \textbf{Verify} or \textbf{Falsify}.

On 5 of the 6 benchmarks, and overall, \tool{} ranks at the top, tying with other verifiers on the ACAS Xu and RESNET benchmarks.
Recall that these benchmarks vary significantly in the number of neurons and parameters, with the ACAS Xu models being modestly sized and the CIFAR models being the largest, and \tool{} is the best on both ends of the scale spectrum.   
\tool{} ranks second on the MNISTFC benchmark to \crown{} both solve the same number of problems, but \crown{} verifies a problem that \tool{} does not leading to its higher score.
The MNIST\_GDVB benchmark varies in size from being comparable to the smallest MNISTFC network to larger than the largest MNISTFC network. Still, a key distinguishing feature of the benchmark is the filtration of \textit{easy} problems.  Whereas MNISTFC includes 23 problems that can be falsified, MNIST\_GDVB has none, yet \tool{} performs better on these harder problems. 

While not a factor in our evaluation, we note that several baseline verifiers require hyperparameter tuning. For example, the runscript of \crown{} for VNN-COMP customizes 10 parameters per \emph{each} benchmark to optimize its performance\footnote{For MNIST\_GDVB, the default configuration of \crown{} performed poorly, so we adopted the configuration used for MNISTFC which gave good results for MNIST\_GDVB.}. 
In contrast, when run with all optimizations enabled, which we recommend based on RQ2's findings, \tool{} has two parameters: the degree of parallelism, $n$, and the number of neurons to attempt to stabilize, $k$.   In these experiments, we fixed these at $k=64$ and $n=4000$ for all benchmarks, which we believe is evidence that developers
can more easily apply \tool{} to new benchmarks while achieving good performance.

\begin{tcolorbox}[left=1pt,right=1pt,top=1pt,bottom=1pt]
\textbf{RQ3 Findings}: \tool{} ranks at the top of a set of baseline DNN verifiers that were shown to be the best performers in a recent DNN verification competition~\cite{muller2022third}.  It performs well on smaller problems like ACAS Xu, where techniques with sophisticated abstract domains like \nnenum{} work well.  It performs well on larger problems like CIFAR2020, where techniques like \nnenum{} fail to solve problems and even highly optimized abstraction-based methods like \crown{} fall short.  It performs well on challenging problems like MNIST\_GDVB, forcing verifiers to analyze the combinatorially sized space of activation patterns to verify problems.
\end{tcolorbox}

\section{Threats to Validity}\label{sec:threats}
Regarding threats to internal validity, we built off the existing code base of \neuralsat{}, thereby leveraging that team's efforts to validate their implementation.  We used assertions in almost every function of our implementation to check the correctness properties flowing from our algorithms, e.g., that lower and upper bounds are properly ordered.  Those assertions were enabled during our rigorous testing process that ran all of the VNN-COMP benchmarks through our implementation, where we confirmed the expected results.

We selected VNN-COMP benchmarks to promote comparability and enhance external validity.  Those benchmarks were developed by other researchers to express verification problems for neural networks, e.g., ACAS Xu is a collision avoidance prediction network for small aerial drones.   Based on our own experience and the experience of the VNN-COMP organizers, who found that some of the VNN-COMP benchmarks were \textit{too easy}, we developed a new benchmark, MNIST\_GDVB.  That benchmark was developed using an approach that guarantees a form of systematic diversity across the networks and specifications that comprise the benchmark.  We plan to continue to push for the development of benchmarks that reflect the challenges of DNN verification, but in this work, we believe our benchmarks are broader and more challenging than prior work.

Regarding construct validity, we used standard metrics, like number of problems solved and VNN-COMP score, that have been widely used~\cite{xu2020systematic,muller2022third}.  This makes comparing our results to prior work easier and allows researchers familiar with the metrics to interpret our results easily.   Moreover, the metrics lead to a natural interpretation that permits answering the research questions, e.g., a verifier that solves more problems has better performance.

\section{Related Work}\label{sec:related} 
Research on DNN verification is extensive and continuously expanding. This section provides an overview of established techniques and their accompanying tool implementations.

\textbf{Constraint-based approaches}, e.g., \reluplex{}~\cite{katz2017reluplex}, and its successor \marabou{}~\cite{katz2019marabou,katz2022reluplex}, \texttt{DLV}~\cite{huang2017safety}, \texttt{Planet}~\cite{ehlers2017formal}, 
and \texttt{MIPVerify}~\cite{tjeng2019evaluating} encode the problem as a constraint-solving task. These techniques transform DNN verification into a constraint problem, solvable using tools like SMT solvers (Planet, DLV) or SAT-based approach with custom simplex and MILP solvers (Reluplex, Marabou).
\textbf{Abstraction-based approaches}, e.g., \texttt{AI\(^2\)}~\cite{gehr2018ai2}, \eran{}~\cite{muller2021scaling,singh2019abstract,singh2018fast} (\texttt{DeepZ}, \texttt{RefineZono}, \texttt{DeepPoly}, \texttt{K-ReLU}), \mnbab{}~\cite{ferrari2022complete}, \reluval{}~\cite{wang2018formal}, \neurify{}~\cite{wang2018efficient}, \verinet{}~\cite{henriksen2020efficient}, \nnv{}~\cite{tran2021robustness}, \nnenum{}~\cite{bak2021nnenum,bak2020improved}, \texttt{CROWN}~\cite{zhang2018efficient}, and \crown{}~\cite{wang2021beta}, leverage abstract domains to tackle scalability. These techniques employ various abstract domains, such as intervals, zonotopes, polytopes, and starsets/imagestars, to improve scalability. To address spurious counterexamples due to overapproximations, these methods often iterate to check counterexamples and refine abstractions.
 \neuralsat{}~\cite{duong2023dpllt} integrates DPLL search with abstraction-based theory solving. 
\texttt{OVAL}~\cite{ovalbab} and \dnnv{}~\cite{shriver2021dnnv} serve as platforms that integrate multiple existing DNN verification tools.
\tool{} extends the \neuralsat{} DPLL(T) approach with neuron stabilization, parallel search, and restart.

Common \textbf{abstract domains} used in DNN verification include intervals~\cite{wang2018formal}, zonotopes~\cite{singh2018fast}, polytopes~\cite{singh2019abstract,xu2020fast}, and starsets/imagestars~\cite{bak2020improved,tran2021robustness}. Notably, top verifiers like \eran{}, \mnbab{}, and \nnenum{} leverage multiple abstract domains to enhance their effectiveness. For instance, \eran{} combines zonotopes and polytopes, while \nnenum{} incorporates polytopes, zonotopes, and imagestars.  
Currently, \tool{} employs polytope abstraction for bound tightening but can also use other abstract domains.

\textbf{Heuristics} and \textbf{optimizations} play a crucial role in the efficiency of SAT solving. Modern SAT/SMT solvers~\cite{moura2008z3,kroening2016decision, barrett2011cvc4}, for instance, benefit from strategies such as VSIDS and DLIS for decision (branching), random restart, and clause shortening or deletion to optimize memory utilization and avoid local maxima. 
Specifically for DNNs, \textbf{neuron stability} serves as a hidden metric for assessing the linearity of neurons with piece-wise linear activation functions. In practice, researchers employ heuristics to apply neuron stability to ReLU during the training of neural networks. For example, the RS Loss approach~\cite{xiao2018training} incorporates regularization techniques to train more stable weights.  The linearity grafting technique~\cite{chen2022linearity} directly replaces ReLU activation functions with linear ones to achieve stability. Both unstructured and structured DNN pruning~\cite{zhangheng2022can} can also help network stabilization. We introduce DPLL(T) parallel search and the concept of neuron stabilization to improve the scalability of DNN verification.

\section{Conclusion and Future Work}\label{sec:conclusion}

As the need for formal analysis increases when more neural networks are being deployed in safety-critical areas, the DNN verification field has received great attention in recent years. In this work, we introduce \tool{}, a ReLU-based DNN verification tool that integrates an advanced DPLL(T) search technique in SAT solving with the concept of neuron stability to significantly reduce the search space of DNN verification.
Our evaluation confirms the effectiveness of \tool{}, which establishes a new state-of-the-art in DNN verifications compared to the performances in the recent DNN verification competition.

We have many opportunities to further improve the performance of \tool{}. For example,  we plan to extend the capability of neuron stabilization in other non-DPLL-based DNN verification techniques and explore new decision heuristics for DPLL(T)-based tools.
Moreover, by using DPLL with implication graphs, \tool{} inherits a native mechanism to verify its own results, e.g., using these graphs and conflicting clauses to obtain resolution graphs/proofs and UNSAT cores as proofs of unsatisfiability~\cite{asin2008efficient,kroening2016decision,zhang2003validating}.


\section{Data Availability}
\tool{} and experimental data are available at: \url{https://github.com/veristable/veristable} 

\bibliographystyle{ACM-Reference-Format}
\bibliography{paper,matt}

\end{document}